\PassOptionsToPackage{numbers,compress}{natbib}
\documentclass{article}

\usepackage[nonanonymous]{paper_style}
\usepackage[utf8]{inputenc}
\usepackage[T1]{fontenc}
\usepackage{hyperref}
\usepackage{url}
\usepackage{booktabs}
\usepackage{amsfonts}
\usepackage{amsmath}
\usepackage{amssymb}
\usepackage{nicefrac}
\usepackage{microtype}
\usepackage{xcolor}
\usepackage{colortbl}
\usepackage{graphicx}
\usepackage{array}
\usepackage{multirow}
\usepackage{tabularx}
\usepackage{placeins}

\usepackage{enumitem}
\newcolumntype{Y}{>{\raggedright\arraybackslash}X}

\newcommand{\pathvu}{PathView-Bench}
\newcommand{\regionfov}{Region-FOV}
\newcommand{\slidefov}{Slide-FOV}

\definecolor{markgreen}{HTML}{00A000}
\definecolor{markred}{HTML}{E60000}
\newcommand{\cmark}{\textcolor{markgreen}{\ensuremath{\boldsymbol{\checkmark}}}}
\newcommand{\xmark}{\textcolor{markred}{\ensuremath{\boldsymbol{\times}}}}
\title{PathView-Bench: Can Multimodal Large Language Models Achieve Fine-grained Multiscale Understanding of Pathology Images?}
\author{
\begin{tabular}{cc}
\begin{tabular}[t]{c}
Zongyi Chen\\
\normalfont National Institute for Data Science\\
\normalfont in Health and Medicine, Xiamen University\\
\normalfont Xiamen, China
\end{tabular}
&
\begin{tabular}[t]{c}
Yu Liang\\
\normalfont Institute of Artificial Intelligence\\
\normalfont Xiamen University, Xiamen, China
\end{tabular}\\[1em]
\begin{tabular}[t]{c}
Jie Lin\\
\normalfont Department of Computer Science\\
\normalfont School of Informatics\\
\normalfont Xiamen University, Xiamen, China
\end{tabular}
&
\begin{tabular}[t]{c}
Liansheng Wang\\
\normalfont Department of Computer Science\\
\normalfont School of Informatics\\
\normalfont Xiamen University, Xiamen, China
\end{tabular}
\end{tabular}
}

\begin{document}

\maketitle

\begin{abstract}
Multimodal large language models (MLLMs) are increasingly used to analyze pathology images. However, dominant multimodal benchmarks in pathology mainly score final diagnostic answers, captions, or reports. These evaluations provide limited insight into whether a model understands the multiscale visual content needed for pathology reasoning and decision-making. We introduce \pathvu, a vision-anchored benchmark for fine-grained and multiscale visual understanding in computational pathology. Built from $23$ public pathology imaging datasets with human-supervised labels and spatial annotations, \pathvu\ evaluates MLLM understanding in two fields of view: \regionfov\ for high-resolution local regions and \slidefov\ for macro whole-slide views. By converting raw annotations into deterministic task targets, \pathvu\ enables programmatic scoring of region localization, visual recognition, quantity estimation, spatial reasoning, and insufficient-context judgment. The benchmark contains $14$ VQA-style tasks, $61,673$ images, and $308,070$ samples across $28$ organs and $7,253,526$ annotations. Evaluating $18$ representative general-purpose, medical-domain, and pathology-oriented MLLMs, we observe substantial limitations even in advanced models on fine-grained visual tasks across multiscale pathology images. \pathvu\ provides a reproducible basis for developing and evaluating pathology MLLMs with explicit multiscale visual understanding.
\end{abstract}

\section{Introduction}

Advances in digital pathology have enabled the digitization and computational analysis of histopathological slides, creating promising opportunities for AI-assisted diagnostics~\citep{madabhushi2016digitalpathology,niazi2019digitalpathology,campanella2019clinicalgrade,clam2021,rrtmil2024,m4mil2025,smmile2025,chief2024,musk2025,gpfm2025}. Recently, multimodal large language models (MLLMs) are increasingly being used for whole-slide analysis. By combining visual understanding with reasoning, these approaches are moving beyond single-task classifiers toward more general and fine-grained pathology assistants~\citep{pathasst2024}, including visual retrieval~\citep{quiltllava2025,cpathomni2024,pathagenticrag2025}, report-oriented interaction~\citep{wsicaption2024,histgen2024,histogpt2025,qcagent2026}, multiscale exploration~\citep{slidechat2025,wsillava2025,cpathagent2025}, navigation~\citep{giant2025,pathologycot2025,pathagent2025}, and cross-FOV visual reasoning~\citep{pathfound2025,pathor12025,pathreasonerr12026,wsiagents2025}.

\begin{figure}[t]
  \centering
  \includegraphics[width=0.98\linewidth]{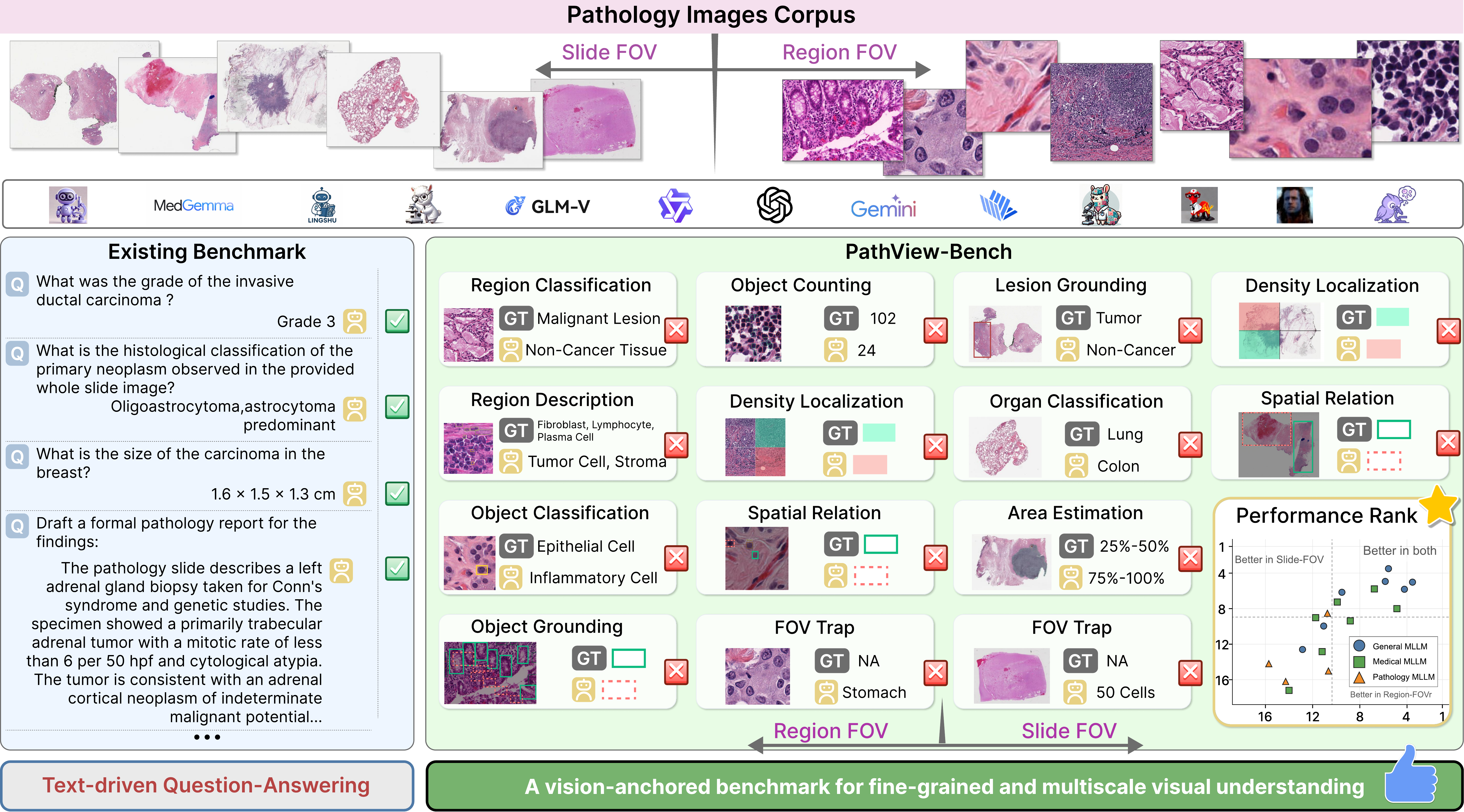}
  \caption{Motivating examples for \pathvu. The benchmark evaluates visual understanding across local-region and whole-slide fields of view through vision-anchored pathology tasks.}
  \label{fig:intro-failure-cases}
  \end{figure}

Existing pathology multimodal datasets and benchmarks mainly assess whether MLLMs can achieve diagnostic correctness~\citep{pathvqa2020,pathmmu2024,wsivqa2024,microbench2024,pathbench2025,pathvg2025}, generate high-quality captions~\citep{quilt1m2023,pathgen2024}, or produce plausible reports across pathology tasks~\citep{wsillava2025,slidechat2025,giant2025}. This text-driven question-answering evaluation paradigm leaves a key question largely unexamined: whether a correct diagnostic answer is supported by fine-grained visual understanding. More concerningly, models might score highly on medical multimodal benchmarks even without image input~\citep{mirage2026,pathmmu2024}. This uncertainty makes it difficult to assess whether the diagnostic output is trustworthy. Moreover, pathology images are extremely large, and local details can differ substantially from global semantic context, yet existing benchmarks often remain limited to a single field of view and therefore provide limited evidence about whether models can understand pathology images across scales. As illustrated in Figure~\ref{fig:intro-failure-cases}, these limitations leave a critical gap for advancing pathology MLLMs:

\textit{\textbf{How can we evaluate their ability to understand fine-grained visual content in multiscale pathology images during reasoning and decision-making?}}

To address this gap, a benchmark is needed to make the visual basis of pathology reasoning measurable. We therefore introduce \pathvu, a vision-anchored benchmark for fine-grained and multiscale visual understanding in computational pathology. \pathvu\ is organized around two fields of view: \regionfov\ for high-resolution local regions and \slidefov\ for macro whole-slide views, reflecting the fact that pathology reasoning depends on both cellular morphology and tissue organization. Across these views, \pathvu\ probes evidence-grounded visual understanding through tasks such as region localization, visual recognition, quantity estimation, spatial reasoning, and judging whether the current view contains sufficient evidence to answer. This shifts evaluation away from final-response plausibility and toward evidence grounded in the multiscale visual context.

To construct the benchmark, we map $23$ publicly available pathology imaging datasets with human-supervised labels and spatial annotations into a shared, auditable benchmark space, where deterministic rules convert the annotations into task targets for programmatic scoring. The resulting benchmark contains $14$ VQA-style tasks, $61,673$ images, and $308,070$ samples across $28$ organs and $7,253,526$ annotations. Across $18$ representative general-purpose, medical-domain, and pathology-oriented MLLMs, \pathvu\ reveals a recurring gap: models are stronger at naming visual categories than at grounding, counting, spatially organizing, or judging the sufficiency of the underlying evidence. Matched-slide analyses further show that correct slide-level diagnostic answers can mask failures in these basic visual operations, and neither larger scale nor pathology-specific training consistently closes the gap.

In summary, the contributions of this work are as follows:
\begin{enumerate}[leftmargin=*,noitemsep,topsep=0pt,parsep=0pt]
\item We introduce a vision-anchored benchmark for fine-grained and multiscale visual understanding in pathology, organized around \regionfov/\slidefov\ and designed to evaluate whether model responses are grounded in observable image content.
\item We define an auditable construction and evaluation protocol that standardizes human-supervised labels and spatial annotations into deterministic task targets with programmatic scoring, without LLM-as-a-judge evaluation.
\item We conduct comprehensive evaluations of $18$ representative general-purpose, medical-domain, and pathology-oriented MLLMs on \pathvu, characterizing their capability boundaries in fine-grained multiscale visual understanding. We further include a PathView-tuned baseline to test whether supervision provides learnable signals for grounding-sensitive behavior.
\end{enumerate}

\section{Related Work}

\paragraph{Multimodal Large Language Models.}
Recent general-purpose MLLMs have broadly advanced multimodal perception, interaction, and reasoning~\citep{gpt4o2024,qwen3vl2025,gemini252025}. This progress has been adapted to medicine through medical image-text pretraining~\citep{medflamingo2023,llavamed2023}, broad biomedical supervision~\citep{medpalm2023}, and domain-specific instruction tuning or medical reasoning alignment~\citep{huatuogptvision2024,lingshu2025}. In pathology, the same trajectory has moved from patch-level dialogue and visual instruction tuning~\citep{pathasst2024,quiltllava2025,cpathomni2024}, to report generation~\citep{wsicaption2024,histgen2024,histogpt2025}, and then toward whole-slide interaction, WSI-level modeling, navigation, and evidence-seeking reasoning~\citep{slidechat2025,wsillava2025,giant2025,pathfound2025,pathor12025,pathreasonerr12026}. These models motivate a stricter evaluation regime: once systems claim richer whole-slide behaviors, they should be judged not only by linguistic fluency but also by whether their outputs remain grounded in visible pathology content.

\paragraph{Pathology MLLM Benchmarks.}
Pathology multimodal evaluation has progressed from patch image question answering and image-text resources to WSI-level interaction. PathVQA~\citep{pathvqa2020} established pathology VQA on patch images, while Quilt-1M~\citep{quilt1m2023} and PathGen-1.6M~\citep{pathgen2024} provided large-scale histopathology image-text pairs for representation learning. Later resources broadened the evaluation scope: PathMMU~\citep{pathmmu2024} and Micro-Bench~\citep{microbench2024} emphasized expert-level or microscopy-oriented local understanding, WSI-VQA~\citep{wsivqa2024}, WSI-Bench~\citep{wsillava2025}, SlideBench~\citep{slidechat2025}, and MultiPathQA~\citep{giant2025} moved toward whole-slide QA and navigation, and PathBench~\citep{pathbench2025} covered both patch and WSI settings. PathVG~\citep{pathvg2025} is closer to explicit visual grounding, but remains focused on region-level localization. Overall as shown in  Table~\ref{tab:benchmark-comparison}, existing resources either emphasize local morphology without global slide context, or evaluate WSI-level answers without requiring grounded intermediate visual analysis. \pathvu\ is designed to cover both \regionfov\ and \slidefov\ while making visual grounding and fine-grained visual-understanding operations central to the evaluation.

\begin{table*}[t]
\caption{Comparison of pathology VQA-style multimodal benchmarks. All-WSI Source indicates whether all images are derived from WSIs. Annotation: LLM = language-model-assisted generation, Exp. = expert annotation or validation, Rule = fixed-rule conversion.}
\label{tab:benchmark-comparison}
\centering
\small
\resizebox{\textwidth}{!}{
\begin{tabular}{lrrcccccc}
\toprule
\textbf{Benchmark} & \textbf{\#Samples} & \textbf{\#Images} & \textbf{\#Task Types} & \textbf{Annotation Method} & \textbf{All-WSI Source} & \textbf{Vision-anchored} & \textbf{Region-FOV} & \textbf{Slide-FOV} \\
\midrule
PathVQA~\citep{pathvqa2020} & 32,799 & 4,998 & 7 & Exp.+Rule & \xmark & \xmark & \cmark & \xmark \\
PathMMU~\citep{pathmmu2024} & 33,428 & 24,067 & 1 & LLM+Rule+Exp. & \xmark & \xmark & \cmark & \xmark \\
WSI-VQA~\citep{wsivqa2024} & 8,672 & 977 & 1 & LLM+Rule & \cmark & \xmark & \xmark & \cmark \\
Micro-Bench~\citep{microbench2024} & 8,485 & 8,385 & 3 & LLM+Exp. & \xmark & \cmark & \cmark & \xmark \\
WSI-Bench~\citep{wsillava2025} & 179,569 & 9,850 & 4 & LLM+Exp. & \cmark & \xmark & \xmark & \cmark \\
PathBench~\citep{pathbench2025} & 33,235 & 12,060 & 2 & LLM+Exp. & \xmark & \xmark & \cmark & \cmark \\
SlideBench~\citep{slidechat2025} & 15,835 & 1,792 & 2 & LLM+Rule+Exp. & \cmark & \xmark & \xmark & \cmark \\
PathVG~\citep{pathvg2025} & 33,500 & 27,610 & 1 & LLM+Exp. & \cmark & \cmark & \cmark & \xmark \\
MultiPathQA~\citep{giant2025} & 934 & 868 & 1 & Exp.+Rule & \cmark & \xmark & \xmark & \cmark \\
PathView (ours) & 308,070 & 61,673 & 14 & Exp.+Rule & \cmark & \cmark & \cmark & \cmark \\
\bottomrule
\end{tabular}
}
\end{table*}

\section{\pathvu\ }

In this section, we describe the design and construction of \pathvu, including its dual-scale task organization, annotation-grounded protocol, and deterministic split and evaluation procedure.

\subsection{Task Design}
WSIs are gigapixel-scale images whose diagnostically relevant information appears at different visual scales~\citep{vanderlaak2021deephistopathology}. In routine pathology review, pathologists assess whole-slide tissue architecture and lesion distribution, while also examining selected high-resolution regions that contain essential morphological details~\citep{pathologycot2025}. \pathvu\ follows this dual-scale reading process by organizing its VQA-style tasks into two fields of view: \slidefov\ tasks for macro whole-slide understanding and \regionfov\ tasks for fine-grained local-region understanding, as shown in Table~\ref{tab:pathview_tasks}.

\begin{table*}[t]
\caption{Overview of the diverse task types, query and answer examples of \pathvu\ }
\label{tab:pathview_tasks}
\centering
\scriptsize
\begin{tabularx}{\textwidth}{@{}l >{\raggedright\arraybackslash}X l@{}}
\toprule
\textbf{Task name} & \textbf{Query example} & \textbf{Answer example} \\
\specialrule{0.4pt}{0pt}{0pt}
\rowcolor{black!8}[0pt][0pt]
\multicolumn{3}{@{}c@{}}{\textbf{Region FOV}} \\
\specialrule{0.4pt}{0pt}{0pt}
Object Grounding &  Locate all \textless{}object\textgreater{}. & [0, 0, 8, 12], ... \\
Object Classification &  Which category best describes the predominant finding in this region \textless{}bbox\textgreater{}? & A. plasma cell \\
Region Classification &  Which category best describes the predominant finding in this image? & A. tumor cell \\
Object Counting &  How many \textless{}object\textgreater{} are present in this image? & 35 \\
Region Description &  Describe what is in the image. & eosinophil, ... \\
Density Localization &  Which quadrant contains the densest distribution of \textless{}object\textgreater{}? & Bottom-right quadrant \\
Spatial Relation &  Local coordinate \textless{}bbox\textgreater{} marks a core region. Which category is  nearest to this region? & B. tumor cell \\
FOV Trap &  Which organ is this image most likely from? & Insufficient information \\
\specialrule{0.4pt}{0pt}{0pt}
\rowcolor{black!8}[0pt][0pt]
\multicolumn{3}{@{}c@{}}{\textbf{Slide FOV}} \\
\specialrule{0.4pt}{0pt}{0pt}
Organ Classification &  Which organ is this image most likely from? & B. spleen \\
Lesion Grounding &  Locate all regions of \textless{}object\textgreater{}. & [50, 138, 87, 227], ... \\
Object Classification &  Which category best describes the predominant finding in this region \textless{}bbox\textgreater{}? & A. malignant lesion \\
Area Estimation &  What percentage of the tissue area in the image is occupied by \textless{}object\textgreater{}? & A. 0\%-25\% \\
Density Localization &  Which quadrant contains the densest distribution of \textless{}object\textgreater{}? & A. Top-left quadrant \\
FOV Trap &  How many \textless{}cell\textgreater{} are present in this image? & Insufficient information \\
\bottomrule
\end{tabularx}
\end{table*}

\subsubsection{Slide-FOV Whole-Slide Understanding}

\slidefov\ tasks evaluate visual understanding from macro whole-slide views. They focus on information that can be assessed at low magnification, including tissue context, lesion location, abnormal burden, and large-scale spatial distribution. These tasks test whether a model can interpret slide-level visual patterns and recognize when the current view is sufficient for the requested inference.

\paragraph{Organ Classification.}
Identifying the organ or tissue source is a basic prerequisite for interpreting a macro view with representative tissue morphology. In this task, the model is asked to infer the most likely organ or tissue source from the selected whole-slide view.

\paragraph{Object Classification.}
Object Classification evaluates whether the model can interpret a specified macro-scale object or ROI. The model is provided with a region on the macro image and asked to classify the predominant pathological or tissue category inside the marked area. Candidate objects are constructed from projected standard annotations.

\paragraph{Lesion Grounding.}
Whole-slide analysis often starts by locating suspicious regions before local inspection. Lesion Grounding asks the model to identify all macro-view regions matching a target lesion or abnormal structure. Multiple independent lesions on the same slide are kept separate to preserve multifocal pathology.

\paragraph{Area Estimation.}
Lesion burden is routinely assessed when estimating tumour extent, necrosis, or other abnormal tissue components. Area Estimation evaluates this ability by asking the model to select the interval that best matches the fraction of visible effective tissue occupied by a target lesion or structure. We compute the fraction over the effective tissue area in the macro image and use four discretized intervals as answer options.

\paragraph{Density Localization.}
Estimating the density of a lesion helps distinguish focal, clustered, and regionally concentrated abnormalities. Density Localization divides the macro image into four quadrants and asks the model to identify the quadrant with the densest target distribution. The target quadrant is determined from both the number of target regions and their occupied area.

This group further includes a \textbf{Slide-FOV trap}, where the model receives a macro WSI view but is asked to count cell-level objects that are not visible at that scale. The correct response is the insufficiency option, not an unsupported count.

\subsubsection{Region-FOV Fine-Grained Histological Understanding}

\regionfov\ tasks evaluate fine-grained visual understanding at the microscopic scale, where relevant visual content appears as cellular morphology, tissue microstructures, annotated object instances, and local spatial relationships. These tasks test whether a model can interpret the histological entities visible in a high-resolution field, rather than relying on slide-level recognition alone.

\paragraph{Region Classification.}
At high magnification, local interpretation often begins with identifying the dominant cellular or tissue component in the field. Region Classification evaluates this ability by asking the model to assign the image to its primary morphology category.

\paragraph{Region Description.}
Pathology image understanding also requires concise descriptions of the components visible in a local field. Region Description asks the model to describe the main objects or structures in the image using standardized terminology. This task does not require a full diagnostic report; instead, it tests whether the model can summarize the key histological components in a local view in a reproducible form.

\paragraph{Object Classification.}
A local histology field can contain multiple annotated objects, including different cell types, tissue structures, or pathological components. Object Classification evaluates whether the model can assign the correct category to a specified instance, rather than only describe the image at the field level.

\paragraph{Object Grounding.}
High-magnification interpretation often depends on accurately localizing key cells, glandular structures, necrotic regions, and other morphology-defined entities. Object Grounding evaluates whether the model can localize these fine-grained pathological entities in a high-resolution local image.

\paragraph{Object Counting.}
Counting cells or small structures is central to many microscopic pathology assessments, including mitotic counts, inflammatory-cell evaluation, positive-cell estimation, and grading or scoring workflows. Object Counting asks the model to report the number of visible target-category instances in a local image, testing whether it can make precise quantitative judgments in dense fields with visually similar or partially overlapping objects.

\paragraph{Density Localization.}
The spatial arrangement of cells or structures in a local field can indicate clustering or heterogeneity. Density Localization asks the model to identify the quadrant with the highest density of the target category.

\paragraph{Spatial Relation.}
Local interpretation can also depend on proximity relationships between histological entities at a comparable scale. Spatial Relation asks the model to identify which candidate category is closest to a specified target instance, using annotated local geometry as supervision.

This group also includes a \textbf{Region-FOV trap}, where the model receives a high-magnification crop with organ-agnostic morphology, such as malignant cells that can appear similar across organs, but is asked to infer the source organ. Because this information is not reliably available from the local crop alone, the correct response is the insufficiency option rather than an unsupported slide-level inference.

\subsection{Standardized Protocol}

\begin{figure}[t]
\centering
\includegraphics[width=0.98\linewidth]{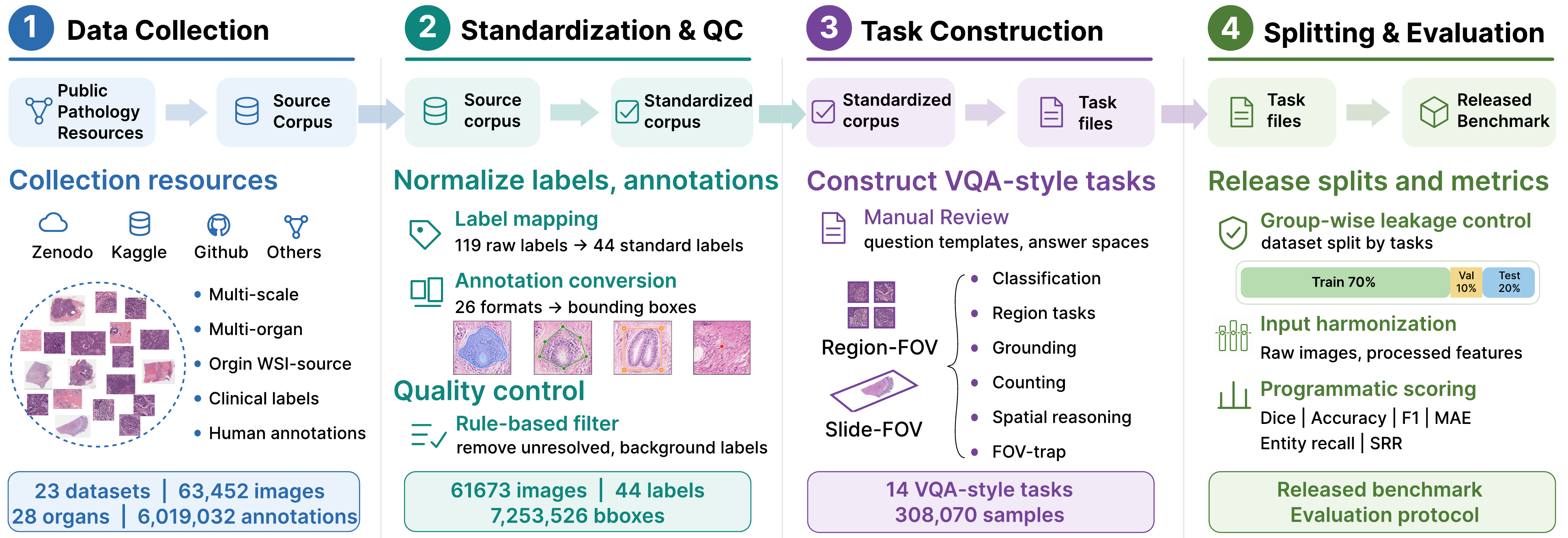}
\caption{Overview of the \pathvu\ standardized protocol pipeline.}
\label{fig:protocol-pipeline}
\end{figure}

\begin{figure}[t]
\centering
\begin{minipage}[c]{0.235\linewidth}
\centering
\parbox[c][\linewidth][c]{\linewidth}{\centering\includegraphics[width=\linewidth,height=\linewidth,keepaspectratio]{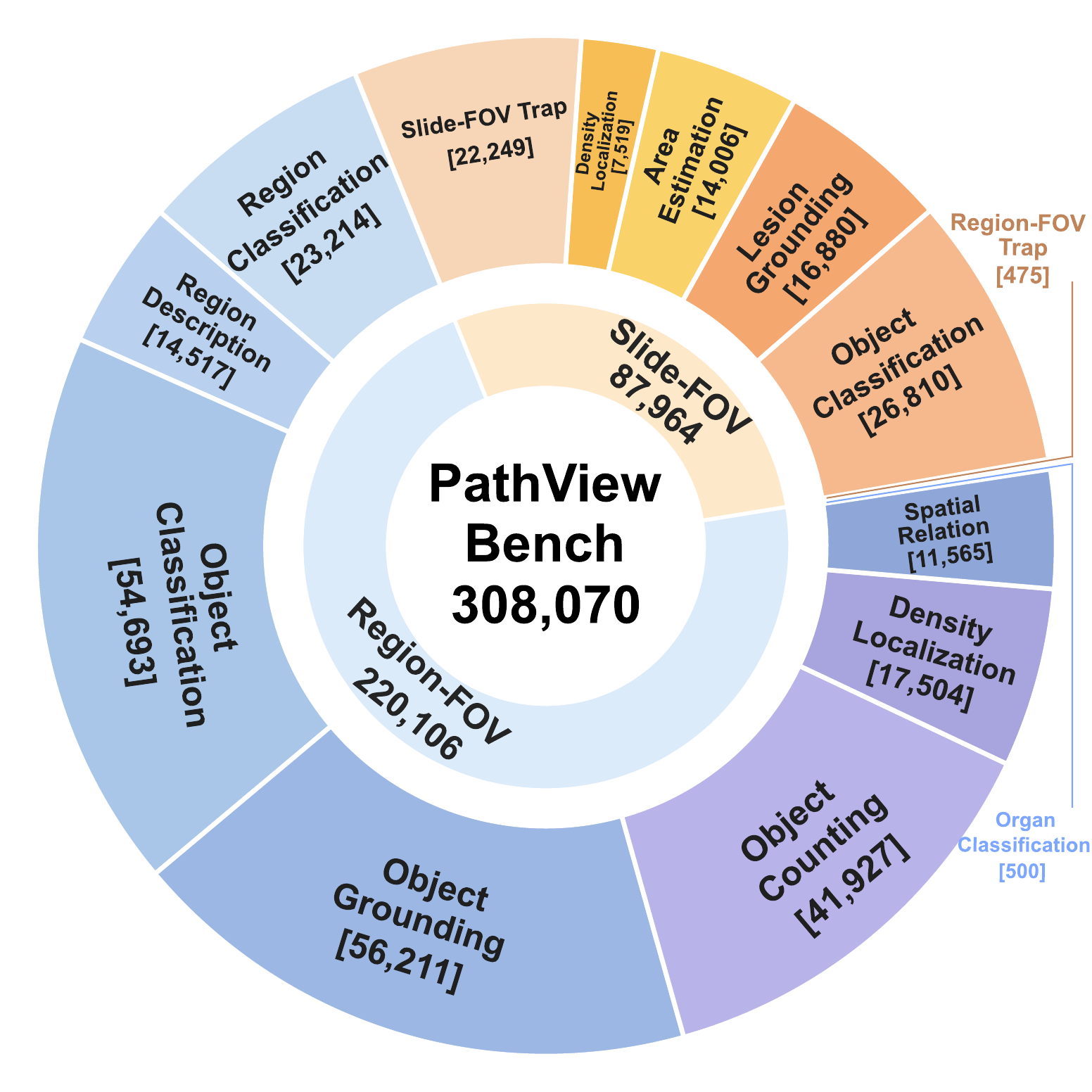}}
\end{minipage}
\hfill
\begin{minipage}[c]{0.235\linewidth}
\centering
\parbox[c][\linewidth][c]{\linewidth}{\centering\includegraphics[width=\linewidth,height=\linewidth,keepaspectratio]{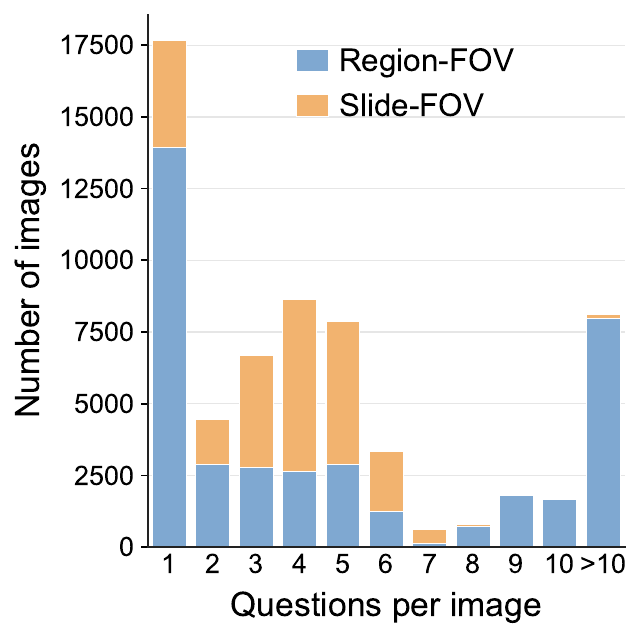}}
\end{minipage}
\hfill
\begin{minipage}[c]{0.235\linewidth}
\centering
\parbox[c][\linewidth][c]{\linewidth}{\centering\includegraphics[width=\linewidth,height=\linewidth,keepaspectratio]{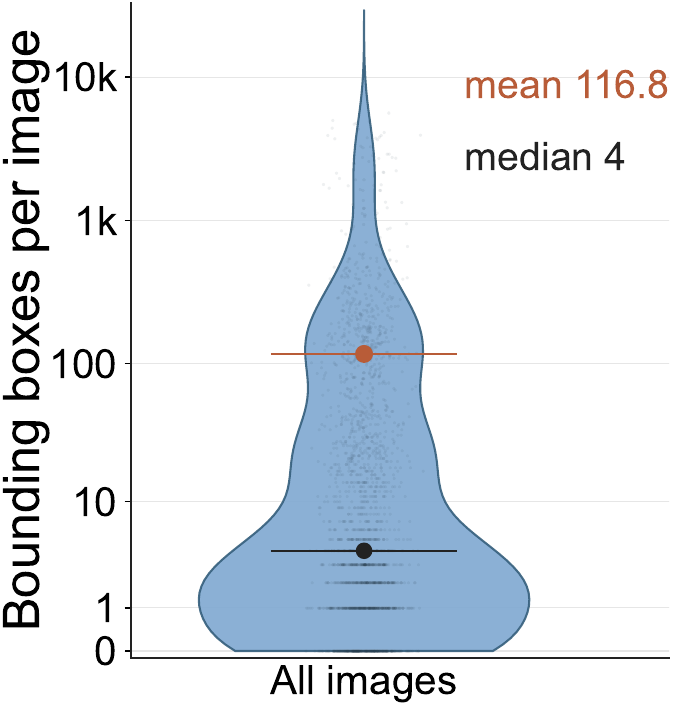}}
\end{minipage}
\hfill
\begin{minipage}[c]{0.235\linewidth}
\centering
\parbox[c][\linewidth][c]{\linewidth}{\centering\includegraphics[width=\linewidth,height=\linewidth,keepaspectratio]{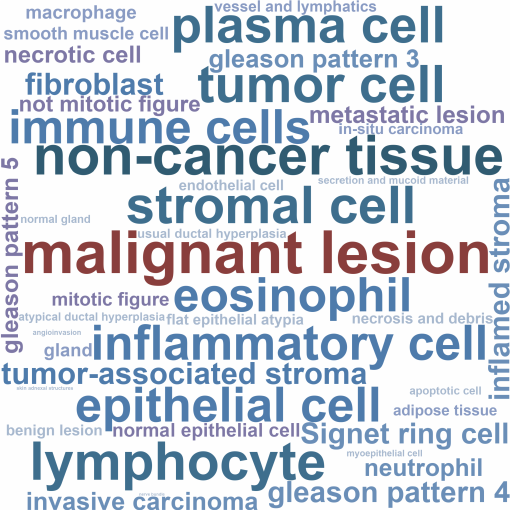}}
\end{minipage}
\caption{Dataset statistics for \pathvu. The four panels summarize the distribution of VQA-style tasks, the number of questions per image, the distribution of spatial annotations per image, and the standardized pathology label vocabulary used to construct the benchmark.}
\label{fig:dataset-statistics}
\end{figure}

To keep benchmark supervision tied to reliable visual content, we treat benchmark construction as an auditable conversion problem rather than automatic data repackaging or model-generated relabeling. Our standardized protocol maps public pathology resources with human-supervised labels and annotations into task targets only when the resulting target is label-valid, observable, and answerable. The protocol consists of four stages: data collection, data standardization and quality control, task construction, and dataset splitting and evaluation. Figure~\ref{fig:protocol-pipeline} summarizes the full protocol, and Figure~\ref{fig:dataset-statistics} reports the resulting dataset statistics.

\subsubsection{Data collection}

We screened public sources, including Zenodo, Kaggle, GitHub, and dataset challenge websites, and retained pathology datasets containing real images with human-supervised labels or annotations. This yielded a source corpus of $23$ datasets, comprising $63,452$ image records and $6,019,032$ raw annotation entries. Patch-level and annotated-region datasets mainly support \regionfov\ tasks, whereas whole-slide datasets support \slidefov\ tasks.

\subsubsection{Data Standardization and Quality Control}

The source corpus differed substantially in label vocabularies and annotation formats. For each dataset, experts reviewed the mapping from raw pathology labels to a shared benchmark vocabulary. We then implemented dataset-specific scripts to convert native spatial annotations into bounding boxes and normalized the resulting boxes to a common coordinate system. Overall, this process consolidated $119$ raw annotation labels into $44$ standard labels and converted $26$ source-specific annotation formats into bounding boxes. Using predefined thresholds and decision rules, we filtered out records with unresolved label mappings, non-pathological background categories, or spatial annotations that were invalid, out of bounds, excessively overlapping, or zero-area. The resulting corpus contains task-ready labels and annotations for downstream task construction.

\subsubsection{Task Construction}

We first defined a standard template for each task, specifying the query format, answer format, candidate answer space, and required supervision. Using these templates, we generated task instances from the standardized corpus according to each dataset's field-of-view scale and available annotations. During generation, we removed samples whose answers could not be uniquely determined, including borderline cases at answer-option thresholds, targets that were invalid or not observable at the current field of view, and cases for which the predefined area, count, or distance criteria could not be computed. We manually reviewed representative converted samples for label consistency, visual observability, and deterministic answerability.

\subsubsection{Dataset Splitting and Evaluation}

We merged the constructed task files by task type while retaining sample-level provenance, including the source dataset, visual input path, task metadata, and split assignment. \pathvu\ uses training, validation, and test splits in a 7:1:2 ratio. To reduce leakage, samples from each source dataset are split using the most specific grouping key available. We provide standard image inputs for image-based models and model-specific feature files for WSI models~\citep{slidechat2025,wsillava2025} that use feature-based inference. Model outputs are scored with task-native deterministic metrics rather than LLM-as-judge evaluation, including Dice for grounding, accuracy for closed-set classification and spatial relation, F1 for interval or quadrant reasoning, MAE for counting, entity recall for description, and SRR for FOV-trap diagnostics.

\begin{table*}[t]
\caption{Main benchmark results on \pathvu. Region-FOV columns: OGD = Object Grounding, RCL = Region Classification, OCL = Object Classification, OCN = Object Counting, RDS = Region Description, DLO = Density Localization, SPR = Spatial Relation. Slide-FOV columns: ORG = Organ Classification, LGD = Lesion Grounding, OCL = Object Classification, ARE = Area Estimation, DLO = Density Localization. OCN is lower better, all other columns are higher better. Best results are bolded and second-best results are underlined. RDS is not assigned to random/frequency baselines because entity recall is not meaningful for frequent-choice description.}
\label{tab:main-results}
\centering
\scriptsize
\resizebox{\textwidth}{!}{
\begin{tabular}{ll*{12}{c}}
\toprule
\multicolumn{2}{c}{} & \multicolumn{7}{c}{Region-FOV} & \multicolumn{5}{c}{Slide-FOV} \\
\cmidrule(lr){3-9}\cmidrule(lr){10-14}
Model & Params & OGD & RCL & OCL & OCN & RDS & DLO & SPR & ORG & LGD & OCL & ARE & DLO \\
\midrule
\multicolumn{14}{l}{\textit{Control baselines}} \\
Random Choice & -- & 0.073 & 0.189 & 0.175 & 482.21 & -- & 0.179 & 0.201 & 0.280 & 0.151 & 0.311 & 0.181 & 0.179 \\
Frequent Choice & -- & 0.000 & 0.247 & 0.213 & 22.23 & -- & 0.102 & 0.260 & 0.340 & 0.005 & 0.465 & 0.103 & 0.105 \\
\midrule
\multicolumn{14}{l}{\textit{General-purpose MLLMs}} \\
Gemini-2.5-Flash & -- & 0.170 & 0.559 & 0.302 & 83.08 & 0.224 & 0.289 & 0.362 & \textbf{0.940} & 0.443 & 0.548 & 0.250 & 0.449 \\
GPT-5-mini & -- & 0.185 & 0.598 & 0.290 & \underline{16.82} & 0.011 & \underline{0.311} & \underline{0.413} & 0.820 & 0.437 & 0.457 & 0.367 & 0.417 \\
InternVL3 & 8B & 0.052 & 0.405 & 0.281 & 28.59 & 0.006 & 0.189 & 0.350 & 0.540 & 0.399 & 0.521 & 0.142 & 0.336 \\
InternVL3.5 & 8B & 0.189 & 0.475 & \underline{0.303} & 19.44 & 0.077 & 0.310 & 0.398 & 0.740 & 0.310 & 0.593 & 0.215 & \underline{0.475} \\
Qwen3-VL & 8B & 0.011 & 0.349 & 0.216 & 21.47 & 0.011 & 0.171 & 0.297 & 0.580 & 0.146 & 0.391 & 0.201 & 0.312 \\
Qwen3-VL & 30B & 0.164 & 0.578 & 0.287 & 19.41 & 0.061 & 0.272 & 0.302 & 0.800 & 0.383 & 0.600 & 0.241 & 0.415 \\
GLM-4.6V-Flash & 10B & \underline{0.201} & 0.348 & 0.275 & 22.95 & 0.006 & 0.250 & 0.264 & 0.530 & \underline{0.552} & 0.548 & 0.219 & 0.447 \\
\midrule
\multicolumn{14}{l}{\textit{Medical-domain MLLMs}} \\
LLaVA-Med v1.5 & 7B & 0.020 & 0.210 & 0.001 & 32.80 & 0.292 & 0.017 & 0.010 & 0.080 & 0.011 & 0.365 & 0.129 & 0.084 \\
HuatuoGPT-V & 7B & 0.066 & 0.352 & 0.190 & 20.17 & 0.147 & 0.230 & 0.327 & 0.630 & 0.151 & \underline{0.603} & 0.246 & 0.409 \\
HuatuoGPT-V & 34B & 0.076 & 0.549 & 0.246 & 37.73 & 0.048 & 0.162 & 0.317 & 0.700 & 0.196 & 0.532 & 0.269 & 0.238 \\
Lingshu & 7B & 0.080 & 0.601 & 0.270 & 19.60 & 0.149 & 0.200 & 0.351 & 0.900 & 0.249 & 0.549 & 0.338 & 0.396 \\
Lingshu & 32B & 0.095 & \underline{0.636} & 0.295 & 18.48 & 0.102 & 0.229 & 0.339 & 0.910 & 0.333 & 0.557 & 0.125 & 0.301 \\
Med Gemma & 27B & 0.129 & 0.520 & 0.275 & 21.57 & 0.006 & 0.255 & 0.319 & 0.490 & 0.305 & 0.419 & \underline{0.448} & 0.360 \\
Med Gemma1.5 & 4B & 0.081 & 0.486 & 0.255 & 31.73 & 0.009 & 0.178 & 0.303 & 0.370 & 0.218 & 0.439 & 0.211 & 0.185 \\
\midrule
\multicolumn{14}{l}{\textit{Pathology-oriented MLLMs}} \\
Quilt-LLaVA & 7B & 0.067 & 0.437 & 0.249 & 59.59 & 0.076 & 0.102 & 0.366 & 0.530 & 0.107 & 0.393 & 0.118 & 0.092 \\
Patho-R1 & 7B & 0.065 & 0.556 & 0.294 & 48.11 & 0.005 & 0.207 & 0.283 & 0.900 & 0.015 & 0.489 & 0.310 & 0.323 \\
SlideChat & $\sim$7B & 0.080 & 0.000 & 0.000 & 24.07 & 0.000 & 0.000 & 0.000 & 0.310 & 0.071 & 0.485 & 0.164 & 0.160 \\
WSI-LLaVA & $\sim$7B & 0.018 & 0.193 & 0.064 & 72.36 & \underline{0.316} & 0.082 & 0.210 & 0.510 & 0.030 & 0.398 & 0.002 & 0.020 \\
\midrule
\multicolumn{14}{l}{\textit{PathView-tuned baseline}} \\
Qwen3-VL & 8B & \textbf{0.323} & \textbf{0.829} & \textbf{0.305} & \textbf{8.32} & \textbf{0.790} & \textbf{0.631} & \textbf{0.673} & \underline{0.920} & \textbf{0.617} & \textbf{0.741} & \textbf{0.693} & \textbf{0.783} \\
\bottomrule
\end{tabular}
}
\end{table*}
\section{Experiment}

\subsection{Models}

We benchmark a diverse collection of MLLMs on \pathvu, including 1) general-purpose models with broad visual-language capabilities, 2) medical-domain models adapted with biomedical or clinical multimodal data, and 3) pathology-oriented models designed for pathology images.

\paragraph{General-purpose MLLMs.}
We evaluate general-purpose MLLMs as open-domain baselines, including Gemini-2.5-Flash~\citep{gemini252025}, GPT-5-mini~\citep{gpt5systemcard2025}, InternVL3 (8B)~\citep{internvl32025}, InternVL3.5 (8B)~\citep{internvl352025}, Qwen3-VL (8B and 30B)~\citep{qwen3vl2025}, and GLM-4.6V-Flash (10B)~\citep{glmv2025}.

\paragraph{Medical-domain MLLMs.}
We evaluate medical-domain MLLMs adapted with biomedical or clinical multimodal data, including LLaVA-Med v1.5 (7B)~\citep{llavamed2023}, HuatuoGPT-V (7B and 34B)~\citep{huatuogptvision2024}, Lingshu (7B and 32B)~\citep{lingshu2025}, Med Gemma (27B)~\citep{medgemma2025}, and Med Gemma1.5 (4B)~\citep{medgemma152026}, to test whether broad medical specialization improves pathology visual understanding.

\paragraph{Pathology-oriented MLLMs.}
We further evaluate pathology-oriented models, including Quilt-LLaVA (7B)~\citep{quiltllava2025}, Patho-R1 (7B)~\citep{pathor12025}, SlideChat~\citep{slidechat2025}, and WSI-LLaVA~\citep{wsillava2025}.

\subsection{Implementation Details}

We evaluate frozen models zero-shot using released checkpoints or API endpoints. For coordinate-sensitive tasks, we follow each model's documented output convention before converting predictions to the unified \pathvu\ coordinate system, e.g., Qwen3-VL~\citep{qwen3vl2025} uses $[0,1000]$-normalized coordinates. We also train a PathView-tuned baseline from Qwen3-VL-8B using only the \pathvu\ training set, with LoRA in ms-swift~\citep{swift2024} for 1 epoch, a global batch size of 16, a learning rate of $5\times10^{-5}$, and 4 NVIDIA A100 GPUs.

\subsection{Main Results}

\begin{figure*}[t]
\centering
\begin{minipage}[c]{0.18\textwidth}
\centering
\phantomsection\label{fig:matched-benchmark-quantitative}
\includegraphics[width=\linewidth]{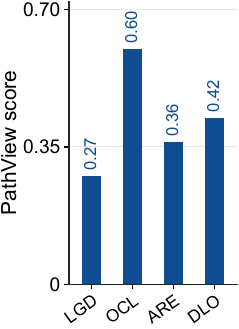}
% \vspace{1.5em}
{}
\end{minipage}
\hfill
\begin{minipage}[c]{0.80\textwidth}
\centering
\phantomsection\label{fig:matched-benchmark-cases}
\includegraphics[width=\linewidth]{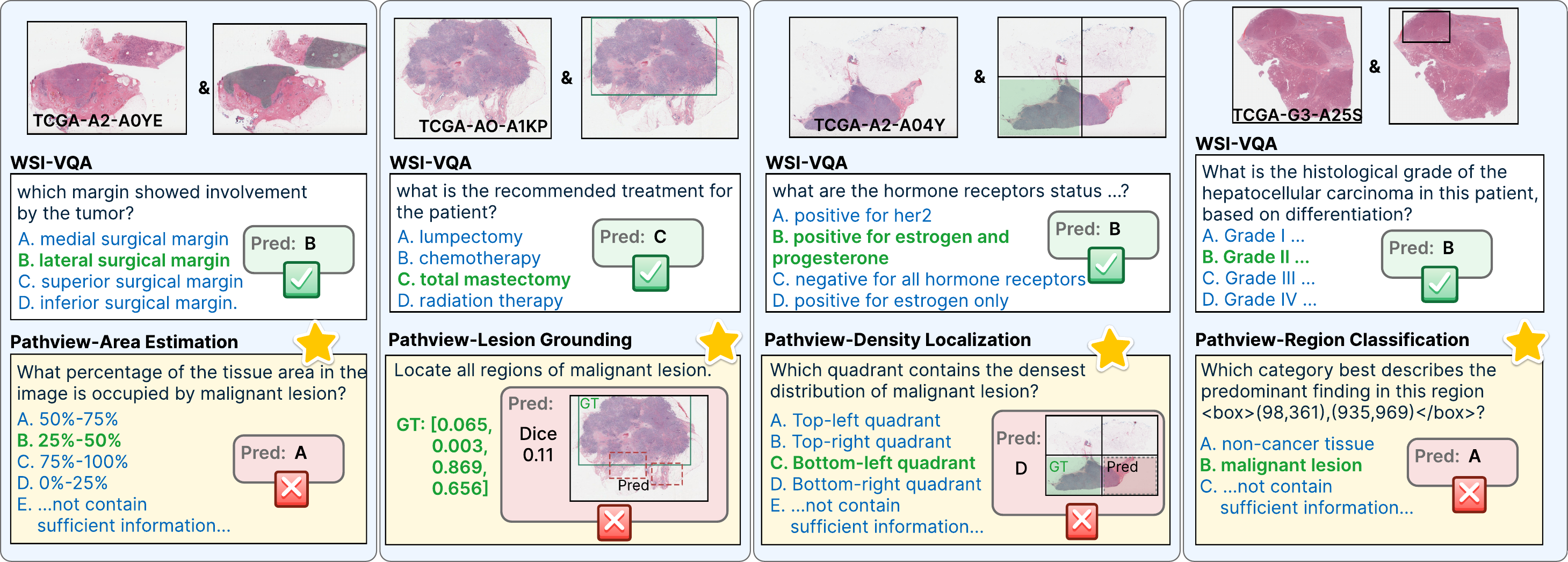}
% \vspace{0.2em}
{}
\end{minipage}
\caption{\textbf{Left}: For each task and model, we first select matched model-slide pairs where the model correctly answers an external SlideBench or WSI-VQA question, and then score the corresponding \pathvu\ visual-understanding task on the same slides. Each bar reports the macro-average across 16 open-source models. \textbf{Right}: shows Lingshu-32B cases in which external benchmark questions are answered correctly but matched \pathvu\ visual-understanding questions fail on the same Slide-FOV slides.}
\label{fig:matched-benchmark}
\end{figure*}

Based on the evaluation results presented in Table~\ref{tab:main-results}, Figure~\ref{fig:refuse-ratio-heatmap}, and Figure~\ref{fig:matched-benchmark}, there are several findings:

\begin{figure*}[t]
\centering
\includegraphics[width=\textwidth]{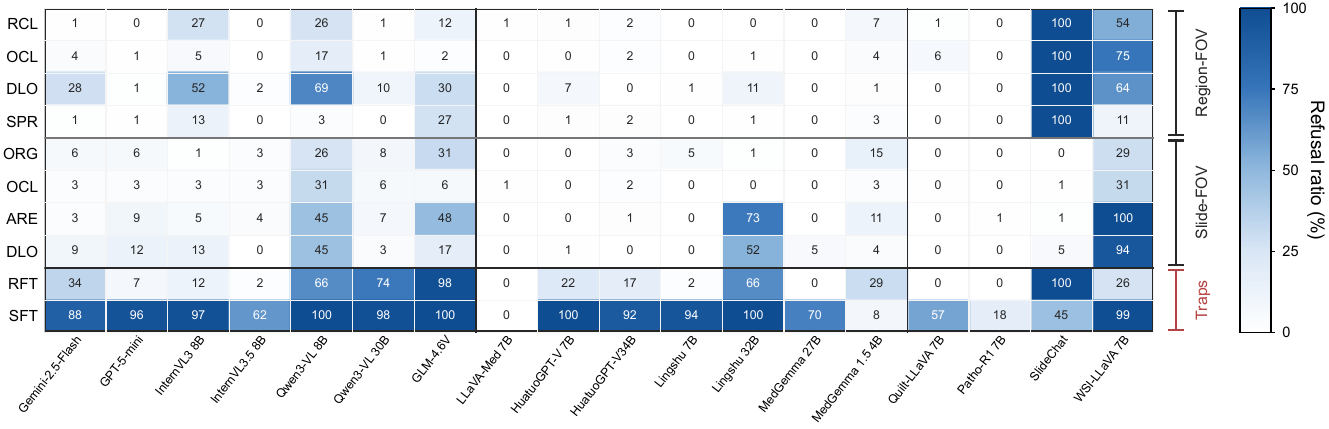}
\caption{Refusal-ratio heatmap for ordinary \pathvu\ tasks and FOV-trap diagnostics. Ordinary tasks measure over-refusal; trap rows measure safe refusal under insufficient field of view. RFT: Region-FOV trap; SFT: Slide-FOV trap.}
\label{fig:refuse-ratio-heatmap}
\end{figure*}

\paragraph{Fine-grained multiscale pathology image understanding remains challenging for MLLMs.}
Our results show that current MLLMs show uneven fine-grained visual understanding across multiscale pathology images. Several frozen models perform well on closed-set recognition, including Gemini-2.5-Flash on Slide-FOV organ classification (0.940) and Lingshu 32B on Region-FOV region classification (0.636), but their performance is lower on tasks that require localization, quantification, or spatial organization. The best frozen model reaches 0.201 Dice on Region-FOV object grounding, and object counting remains difficult, with the best frozen MAE at 16.82. The PathView-tuned baseline shows that \pathvu-derived targets provide useful supervision signals for grounding-sensitive behavior. 

\paragraph{Domain specialization and model scale do not guarantee stronger performance.}
Medical- and pathology-specialized models do not consistently outperform general-purpose MLLMs on \pathvu, suggesting that domain-knowledge training must be balanced with basic visual grounding and spatial reasoning ability. Scaling shows a similarly uneven pattern across model families and fields of view. Separating \regionfov\ and \slidefov\ results exposes these scale- and view-dependent weaknesses. For pathology-oriented models, weaker spatial scores may partly reflect slide-feature interfaces that preserve global semantics but lose pixel-level coordinate relationships. Improving coordinate-aware grounding would help extend these models to evidence navigation, lesion localization, burden estimation, and interactive review.

\paragraph{Models show inconsistent caution under observability limits.}
We analyze refusal behavior because unsupported answers can be unsafe in pathology, while excessive refusal reduces utility on answerable images. As shown in Figure~\ref{fig:refuse-ratio-heatmap}, ordinary tasks have a 12.4\% average refusal rate, whereas safe-refusal rates are 30.9\% for Region-FOV traps and 73.6\% for Slide-FOV traps. Errors are asymmetric: SlideChat, WSI-LLaVA, and Qwen3-VL 8B over-refuse answerable tasks (50.8\%, 57.2\%, and 32.8\%), while LLaVA-Med, Patho-R1, and MedGemma 1.5 4B often answer unsupported trap questions. The main limitation is not simply that models refuse too often or too rarely; rather, they fail to calibrate refusal to the visual evidence available at the current field of view.

\paragraph{Correct high-level answers do not imply correct basic visual understanding.}
To test whether high-level WSI QA performance transfers to basic visual operations, we matched \pathvu\ Slide-FOV samples to slides that also appear in SlideBench~\citep{slidechat2025} or WSI-VQA~\citep{wsivqa2024}. For each model and task, we conditioned the analysis on external success by retaining only matched slides on which the same model correctly answered at external benchmark question, and then scored the corresponding \pathvu\ task on those slides. The conditional \pathvu\ scores remained limited for tasks requiring localization or quantification: 0.274 for lesion grounding, 0.360 for area estimation, and 0.422 for density localization (Figure~\ref{fig:matched-benchmark}). The object-classification subset is less affected because most matched samples ask only for binary cancer-region recognition. Even after conditioning on externally correct high-level answers, the qualitative pairs in Figure~\ref{fig:matched-benchmark}b show failures in basic visual interpretation on the same Slide-FOV slides. Overall, this matched-slide analysis suggests that correct high-level recognition does not reliably imply fine-grained localization, burden estimation, or spatial reasoning.

\section{Limitations and Future Work}

\pathvu\ is built from publicly available pathology datasets and is intended as a vision-centered evaluation resource rather than a complete clinical benchmark. It tests whether pathology MLLMs can ground structured answers in visible image content, but it does not assess full diagnostic reporting quality, clinical deployment readiness, or human-pathologist equivalence. In particular, \pathvu\ does not include private longitudinal records, multi-timepoint diagnostic workflows, or follow-up information. Future extensions will incorporate clinical workflow data, guideline-driven diagnostic observations, and pathologists' visual decision trajectories while preserving \pathvu\ as an auditable benchmark for multiscale visual understanding.

\section{Conclusion}

We introduced \pathvu, a vision-centered benchmark for evaluating pathology MLLMs across region and slide fields of view. Built from standardized public datasets, it uses task-ready labels, geometry, prompts, and deterministic metrics to test localization, classification, quantification, spatial reasoning, and refusal under insufficient visual context. Across models, domain specialization, scale, and high-level WSI QA performance do not reliably translate into fine-grained visual grounding. The PathView-tuned baseline further indicates that structured \pathvu-derived targets provide learnable supervision signals under the benchmark protocol. These findings argue for evidence-grounded evaluation, rather than answer plausibility alone, and for models that preserve spatial evidence across fields of view.

{\small
\bibliographystyle{unsrtnat}
\bibliography{references}
}

\end{document}